\documentclass[letterpaper, 10 pt, conference]{ieeeconf}  

\IEEEoverridecommandlockouts                              

\usepackage{cite}
\usepackage{amsmath,amssymb,amsfonts}
\usepackage{algorithmic}
\usepackage{hyperref}
\usepackage{graphicx}
\usepackage{textcomp}
\usepackage{xcolor}

\newcommand{\projectwebsite}{\url{https://gmargo11.github.io/softmimic}}

\newcommand{\qref}{q_{\text{ref}}}
\newcommand{\qaug}{q_{\text{aug}}}

\newcommand{\Rref}[1]{R_{#1,\text{ref}}}
\newcommand{\pref}[1]{\mathbf{p}_{#1,\text{ref}}}

\newcommand{\Rdes}[1]{R_{#1,\text{des}}}
\newcommand{\pdes}[1]{\mathbf{p}_{#1,\text{des}}}

\newcommand{\wi}{\mathbf{w}_i}
\newcommand{\Fi}{\mathbf{F}_i}
\newcommand{\taui}{\boldsymbol{\tau}_i}
\newcommand{\Kcmd}{\mathbf{K}_{\text{cmd}}}

\newcommand{\Kcmdt}{K^t_{\text{cmd}}}
\newcommand{\Kcmdr}{K^r_{\text{cmd}}}
\newcommand{\Kenvt}{K^t_{\text{env}}}
\newcommand{\Kenvr}{K^r_{\text{env}}}

\usepackage[english]{babel}
\usepackage[T1]{fontenc}

\usepackage[nolist]{acronym}
\usepackage{amsmath}
\usepackage[font={small}]{caption}
\usepackage{subcaption}
\usepackage[export]{adjustbox}
\usepackage{amssymb}
\usepackage{bm}
\usepackage{booktabs}
\usepackage{booktabs}
\usepackage{balance}
\usepackage{cite}
\usepackage{color}
\usepackage{dsfont}
\usepackage{esvect}
\usepackage{float}
\usepackage{graphicx}
\usepackage{hyperref}
\usepackage{import}
\usepackage{mathtools}
\usepackage{multirow}
\usepackage{multicol}
\usepackage{multirow}
\usepackage{flushend}
\usepackage{outline}
\usepackage{paralist}
\usepackage{siunitx}
\usepackage{stackengine}
\usepackage{stmaryrd}
\usepackage{systeme}
\usepackage{soul}
\usepackage{url}
\usepackage{tikz}
\usetikzlibrary{tikzmark}
\usetikzlibrary{calc}
\usepackage[normalem]{ulem}

\title{\LARGE \bf
\textit{SoftMimic}: Learning Compliant Whole-body Control from Examples
}

\author{Gabriel B. Margolis$^*$ \quad Michelle Wang$^*$ \quad Nolan Fey \quad Pulkit Agrawal
\thanks{All authors are with the Improbable AI Lab,
        Massachusetts Institute of Technology, USA. Correspondence to:
        {\tt\footnotesize \{gmargo, wangmj\}@mit.edu}}
\thanks{$^*$ indicates co-first authors.}
\thanks{Website: \projectwebsite}
}

\begin{document}

\let\oldtwocolumn\twocolumn
\renewcommand\twocolumn[1][]{%
    \oldtwocolumn[{#1}{
    \begin{flushleft}
           \centering
    \vspace{-0.5cm}
    \includegraphics[width=1.0\textwidth]{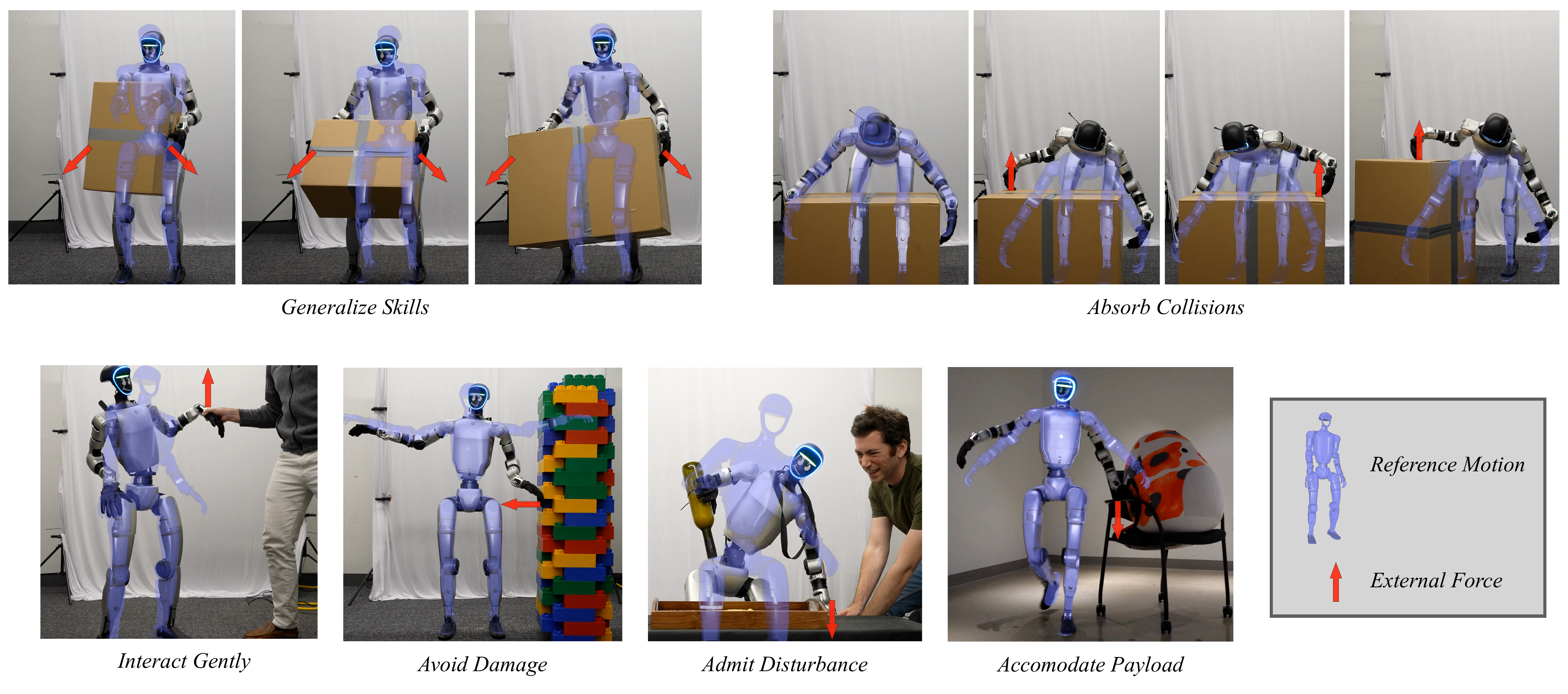}
    \captionof{figure}{
    \textbf{SoftMimic for Compliant Motion Tracking.}  We train humanoid policies that compliantly respond to external forces while tracking a reference motion. The desired force-displacement relationship is modulated by a `stiffness' input at deployment time, and a single policy learns to realize a wide range of stiffnesses. In diverse real-world experiments, SoftMimic benefits generalization and safety. In the images, the reference motion is visualized in blue, and the approximate external force on the robot is illustrated by the red arrows. 
    }\label{fig:setup}
    \end{flushleft}
    }]
}

\maketitle
\thispagestyle{empty}
\pagestyle{empty}

\begin{abstract}

We introduce \textit{SoftMimic}, a framework for learning compliant whole-body control policies for humanoid robots from example motions.  Imitating human motions with reinforcement learning allows humanoids to quickly learn new skills, but existing methods incentivize stiff control that aggressively corrects deviations from a reference motion, leading to brittle and unsafe behavior when the robot encounters unexpected contacts. In contrast, SoftMimic enables robots to respond compliantly to external forces while maintaining balance and posture. Our approach leverages an inverse kinematics solver to generate an augmented dataset of feasible compliant motions, which we use to train a reinforcement learning policy. By rewarding the policy for matching compliant responses rather than rigidly tracking the reference motion, SoftMimic learns to absorb disturbances and generalize to varied tasks from a single motion clip. We validate our method through simulations and real-world experiments, demonstrating safe and effective interaction with the environment.

\end{abstract}

\section{INTRODUCTION}

A major goal in humanoid robotics is to build agents capable of performing a vast range of tasks humans execute in everyday environments. A promising avenue towards this goal is to leverage large-scale human motion capture data, enabling robots to learn human-like behaviors through imitation \cite{peng2018deepmimic}. Recent work has successfully trained policies for tracking single motions, diverse motion datasets, and even real-time teleoperation on humanoid hardware \cite{xie2025kungfubot, cheng2024expressive, he2024learning, chen2025gmt, liao2025beyondmimic}. These methods produce impressive, dynamic motions.

\begin{figure*}[t]
    \centering
    \vspace{0.8em}
    \includegraphics[clip,trim=2cm 0cm 15cm 0cm,width=1.0\linewidth]{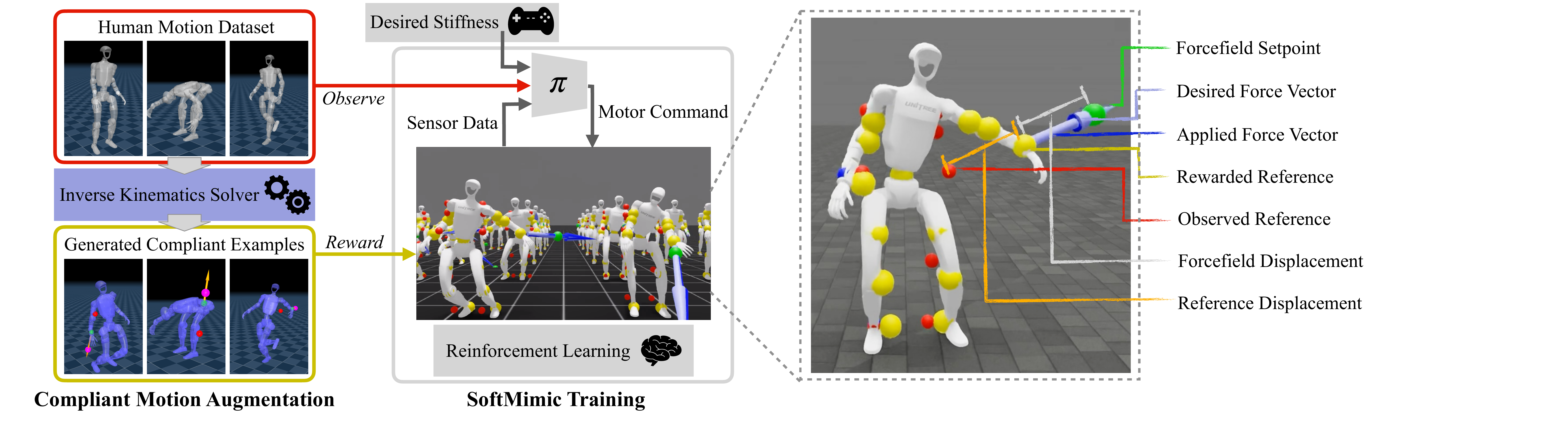}
    \caption{ \textbf{Soft Whole-body Control via Compliant Motion Augmentation.}
    \textbf{\textit{Left}}: Given an original reference motion ($q_\text{ref}$) and a specified interaction (external wrench $W_\text{ext}$ and stiffness $K_\text{robot}$), our offline data generation stage uses an 
    IK solver to generate a kinematically feasible and stylistically consistent compliant motion ($q_\text{aug}$). \textbf{\textit{Right}}: During online training, a policy learns to reproduce this behavior. It observes the robot's proprioceptive state and the original reference ($q_\text{ref}$), but is rewarded for matching the augmented target ($q_\text{aug}$). This forces the policy to implicitly infer the external wrench and react appropriately, resulting in a robot that can controllably comply with generalized unanticipated perturbations. In the graphic, we only annotate translational forces and displacements for ease of interpretation, but the analogous rotational quantities are also simulated.
    }
    \label{fig:system_diagram}
\end{figure*}

However, motion tracking is usually insufficient for safe and effective deployment in the real world, where sensing uncertainty and frequent, unplanned physical (i.e., contact-rich) interactions are commonplace. Policies trained to rigidly track a reference motion treat any deviation in the robot's motion as an error to be corrected aggressively. Consequently, when the robot makes an unexpected contact, such as brushing against a table, misjudging an object's location, or interacting with a person, the controller attempts to correct the motion error caused by the contact with large, uncontrolled forces, resulting in brittle and potentially dangerous behavior. This lack of compliance is also a fundamental barrier to deploying humanoids alongside people, leading to the current state of humanoids operating in isolation.

To address these shortcomings and pave the path for real-world humanoid deployment, we propose a framework for \textit{compliant} whole-body motion tracking called \textit{SoftMimic}. The objective of \textit{SoftMimic} is not to blindly minimize tracking error, but to controllably \textit{depart} from the reference motion in response to external forces according to a user-specified stiffness. A lower stiffness setting allows the robot to comply more and thereby deviate more from the reference trajectory, given the same force disturbance. Achieving compliant behavior on a high-DoF humanoid is challenging because complying with a force on a single limb requires coordinated, full-body adjustments to maintain balance and preserve the overall posture and style of the motion.

Directly learning compliant behavior with reinforcement learning (RL) poses significant exploration challenges, as a stiff, non-compliant policy is often a strong local optimum. 
In scenarios where the robot must comply substantially, reward terms like tracking keypoints and joint angles, which typically reinforce each other, can come into conflict, making it difficult to balance them across a large potential solution space~\cite{sentis2004task}.
Furthermore, many desired compliant responses are kinematically or dynamically infeasible depending on the robot's configuration, resulting in training for impossible tasks, which hinders learning \cite{margolis2022rapid}. 

To overcome these challenges, we adopt a 
learning-from-examples 
strategy. Instead of asking RL to discover compliant behaviors through intricate reward shaping, we first generate a dataset of kinematic references for compliant behaviors. 
We use an inverse kinematics (IK) solver to author a large-scale dataset of feasible and stylistically-consistent compliant trajectories for a wide range of interaction scenarios. This offline process allows us to filter out impossible tasks and precisely define the desired whole-body coordination. We then train a policy using RL, where the agent observes the robot's state and the original, non-compliant reference motion, but is rewarded for tracking the corresponding pre-computed compliant trajectory from our augmented dataset. This formulation forces the policy to learn to infer the external forces from proprioceptive sensing 
and react with the demonstrated compliant behavior.

Our experiments demonstrate that this approach yields a policy capable of tracking reference motions while exhibiting predictable, controllable compliance. Our compliant controller is more robust to disturbances, can generalize a single motion clip to handle variations in a task (e.g., picking up boxes of different sizes), and safely manages unexpected collisions. Crucially, these benefits are achieved while preserving good motion tracking performance in the absence of external forces. 
We validate our framework in simulation and on a real Unitree G1 humanoid robot.

\section{RELATED WORK}

\subsection{Learning Humanoid Whole-Body Control}

The convergence of recent progress in articulated rigid-body simulation \cite{isaacGym, todorov2012mujoco, zakka2025mujoco}, sim-to-real transfer techniques \cite{peng2018sim, hwangbo2019learning, margolis2022walktheseways}, and advanced legged hardware \cite{hutter2016anymal, wensing2017proprioceptive, katz2019mini, unitree2025website}, combined with reinforcement learning paradigms like DeepMimic \cite{schulman2017proximal, peng2018deepmimic}, has enabled impressive performance in humanoid robot motion imitation \cite{chen2025gmt, xie2025kungfubot, liao2025beyondmimic} and real-time teleoperation \cite{cheng2024expressive, he2024learning, li2025amo,ze2025twist}. 
Building on these components, some works use such whole-body controllers to teleoperate new tasks and train visuomotor policies on the resulting demonstrations, establishing a mapping from image observations to whole-body motion references
\cite{fu2024humanplus, ben2025homie, fu2024mobile, fang2025dexop}.

During imitation or teleoperation, a common scenario is that the robot contacts an object while the teleoperator or motion reference does not. Light objects may be pushed out of the way or lifted, but heavier objects or fixtures may impede the robot and inhibit it from matching the reference. A natural question is what posture the robot should adopt when it is forced away from the reference motion, and how much force the robot should exert against the environment when attempting to reduce tracking error. 
Traditional factory arms that are purely position-controlled are extremely stiff, and consequently may damage themselves or the environment with large and unpredictable forces when impeded, making them dangerous and brittle to small environmental variations. Modern quasi-direct-drive (QDD) actuators support torque sensing and control, which allows them to realize different stiffnesses through software emulation. A prevalent strategy within learning-based whole-body control frameworks is for a neural network policy to actuate the robot's QDD motors by sending target setpoints to a PD controller in each joint. 
Such policies can modulate position targets to intentionally incur position error, regulating applied forces during dynamic maneuvers \cite{hwangbo2019learning}. Furthermore, PD gains can be tuned to shape the torque and position distributions excited by Gaussian policy exploration \cite{esser2024action}. 
A natural misconception might be that lower gains or even direct torque control will always result in a compliant or `soft' robot policy. In reality, as we show, the stiffness of a policy's interactions is dictated foremost by its high-level incentives, i.e. its reward function and training data.
We also find neural network policies trained with constant low-level gains are capable of representing a wide range of stiff and compliant behaviors in task-space.

Works that combine motion tracking rewards with random external forces or pushes instruct the robot to follow the same trajectory regardless of the interaction force~\cite{fu2023deep, ze2025twist, zhang2025falcon}. This encourages the policy to apply arbitrary resistive forces to maintain minimal tracking error, essentially acting as stiff as possible.

\subsection{Analytical Approaches to Compliant Control}

Hybrid position/force control \cite{raibert1981hybrid} and task-space impedance–admittance control \cite{hogan1985impedance} are longstanding formulations for compliant manipulation in robotics, which prescribe a motion–wrench relationship (e.g., a virtual mass–spring–damper at the end effector). In one line of work, this relationship is implemented so that the closed loop is \emph{passive} at the interaction ports, emphasizing robustness and safety in contact; in another, inverse kinematics/dynamics (IK/ID) are used to synthesize joint trajectories/torques that realize desired task motions and apparent stiffness given a model. Extending these ideas from fixed-base robot arms to humanoids requires incorporating models of a floating base, intermittent contacts, and the need to coordinate interaction objectives with posture and balance. The operational-space formulation \cite{khatib2003unified} provides an IK/ID approach to control the robot while balancing multiple tasks such as force interaction and posture control. Whole-body operational-space control extended this to floating-base systems under contact constraints, organizing interaction tasks alongside posture and balance through contact-consistent projections \cite{sentis2004task,sentis2005synthesis,sentis2006whole,sentis2010compliant}. Control methods based on passivity have also been developed to balance the robot while ensuring compliant interaction of the hands and feet 
\cite{albu2007unified, hyon2007full, ott2006humanoid, ott2013kinesthetic, henze2015approach, garcia2019integration, abi2019torque}. These analytical approaches have yielded impressive demonstrations of precise force sensing, back-drivability, and safe physical interaction, most notably on the DLR Torque-controlled humanoid Robot (TORO) \cite{englsberger2014overview}.

Drawing inspiration from the above literature, we develop a compliant approach to RL motion imitation (SoftMimic) which incorporates an explicit task-space interaction law. Our approach adopts the classical goal of making the robot behave like a spring in response to generalized disturbances, but replaces hand-engineered controllers with a learned policy trained on procedurally generated compliant trajectories. The compliant trajectory generation is based on simple kinematic heuristics whereas the RL training stage accounts for the full dynamics model. The policy observes proprioception and the original (non-compliant) reference, and is rewarded for reproducing the authored compliant deviation. This allows user-specified stiffness to be realized while maintaining high-fidelity whole-body motion tracking.

\begin{figure}[t]
    \centering
    \vspace{0.5em}
    \includegraphics[width=0.7\columnwidth]{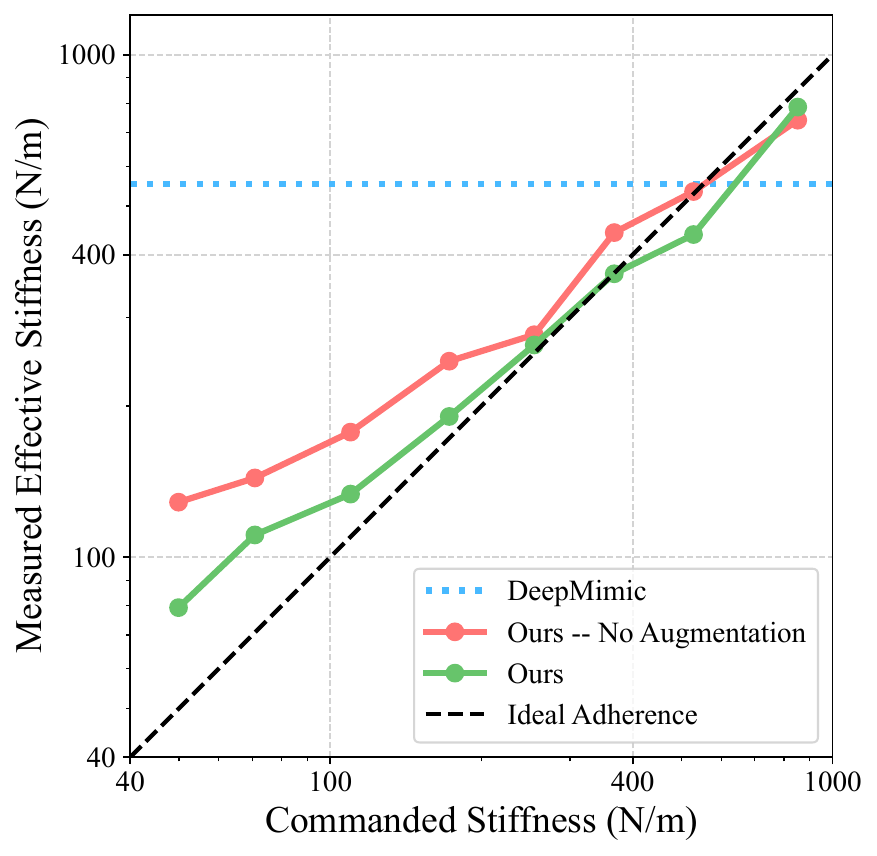}
    \caption{\textbf{Stiffness adherence.}
    The humanoid’s effective translational stiffness tracks the commanded stiffness over a wide range. 
    We apply external forces to the hands of a standing robot in simulation and report the median force–displacement ratio for a single \textit{SoftMimic} policy across stiffness commands (log–log scale). 
    A stiff motion-tracking baseline maintains an approximately constant effective stiffness of \SI{550}{\newton\per\meter} under identical conditions. 
    Data augmentation further improves adherence, especially at low stiffness (Fig.~\ref{fig:aug_benefit}).}
    \label{fig:stiffness_adherence}
\end{figure}

\subsection{Data-Driven Compliant Control}

Recently, reinforcement learning has been used to directly learn compliant behaviors. Initial explorations, such as Deep Compliant Control \cite{lee2022deep}, demonstrated success with simulated characters but relied on perfect state information, sidestepping the real-world challenges of force estimation and model uncertainty. Portela et al. \cite{portela2024learning} made a key step toward hardware applications by showing that an end-to-end policy trained in simulation can learn to apply accurate task-space forces on a real legged manipulator using only proprioceptive actuators, and demonstrated that this facilitates impedance control of the robot's end-effector.
Other work has trained locomotion policies to directly mimic a specific dynamic model, such as a spring-mass-damper template \cite{hartmann2024deep}, a concept extended by FACET \cite{xu2025facet} to various embodiments. UniFP \cite{zhi2025learning} demonstrated that explicit force information obtained from such policies can benefit imitation learning for downstream tasks. These prior approaches focus on controlling the force interaction while satisfying a simple locomotion task. Especially in the case of humanoid robots learning from human teleoperation and/or demonstrations, it is often critical to reconcile interaction objectives with whole-body motion tracking in order to complete a task.
A unified framework that combines wide-range impedance control with high-fidelity motion mimicry on real hardware remains an open challenge. Our work addresses this gap by training a single policy to imitate reference motions while achieving a user-specified stiffness, enabling both soft compliance and stiff resistance (Figure \ref{fig:stiffness_adherence})

\section{Method}
\label{sec:method}

Our goal is to train a policy that enables a humanoid robot to track a whole-body reference motion while compliantly responding to external forces with a user-specified stiffness. A naive approach could involve a standard motion imitation setup \cite{peng2018deepmimic} with an additional task reward for compliant responses. 
Directly optimizing this objective with RL is not ideal for several reasons. 
First, exploration is brittle: a purely stiff tracker is a strong local optimum that suppresses compliant responses when these rewards are in conflict. Second, the humanoid’s large postural null-space makes reward design--balancing interaction forces with whole-body style and stability--nontrivial. 
Third, the robot's feasible compliance is highly dependent on its configuration due to kinematic constraints and sensing limitations, yielding many tasks infeasible. Fourth, the desired deviation from reference motions is incompatible with the use of \textit{early termination} and \textit{reference state initialization} strategies commonly used to stabilize and accelerate training.

Our solution to these problems is to generate an \textit{augmented} dataset with reference motions that specify how the robot should comply to different external forces. This sets up a motion tracking problem where the robot observes the original motion target but is rewarded for inferring the force interaction and matching the applicable augmented target. We show that this approach enables fine-grained control of the compliant response. A key challenge is how to generate a dataset of feasible complying motions that preserve desired components of the original motion style. In this work, we use differential inverse kinematics to ensure kinematically feasible and stylistically desirable reference motions, and an analysis of force and position sensing noise to specify feasible force/compliance tasks.

\subsection{Compliant Motion Tracking (CMT)}

Given an original reference configuration $\qref(t)$ and an external wrench on link $i$, $\wi(t)=(\Fi(t),\taui(t))$, with a commanded robot stiffness
$\Kcmd=\mathrm{diag}(\Kcmdt \mathbf{I}_3,\ \Kcmdr \mathbf{I}_3)$,
we define the \emph{desired compliant target pose} for link $i$ relative to the reference:
\[
  \pdes{i}=\pref{i}+\frac{1}{\Kcmdt}\,\Fi,\qquad
  \Rdes{i}=\Rref{i}\exp\!\big([\taui/\Kcmdr]_\times\big).
\]
Let $T_i(q)=(R_i(q),\mathbf{p}_i(q))$ denote the pose of link $i$. The instantaneous IK objective is
\[
  \hat{q}=\arg\min_q d(q,\qref)\quad
  \text{s.t.}\ \ \mathbf{p}_i(q)\approx \pdes{i},\ \ R_i(q)\approx \Rdes{i}.
\]
This dictates that link $i$ behaves like a spring with stiffness $(\Kcmdt,\Kcmdr)$ around the reference, while the rest of the posture stays as close as possible to $\qref$ under the distance metric $d$. 
When the robot's stiffness is low or the external force is large, the optimal configuration $\hat{q}$ can deviate significantly from the reference $q_{\text{ref}}$. In such cases, the choice of the metric $d$ (e.g., a simple joint-space error like $\| q - q_{\text{ref}} \|^2$ versus a task-space error on other keypoints) has a large impact on the resulting behavior. This contrasts with typical motion tracking scenarios where the optimal solution remains close to the reference, and all errors are near zero. In this work, we choose to define $d$ using a mixture of keypoint error, joint position error, foot placement consistency, and center-of-pressure maintenance as described in Section \ref{sec:data_shaping}.

\begin{figure}[t]
    \centering
    \vspace{0.3em}
    \includegraphics[width=1.0\columnwidth]{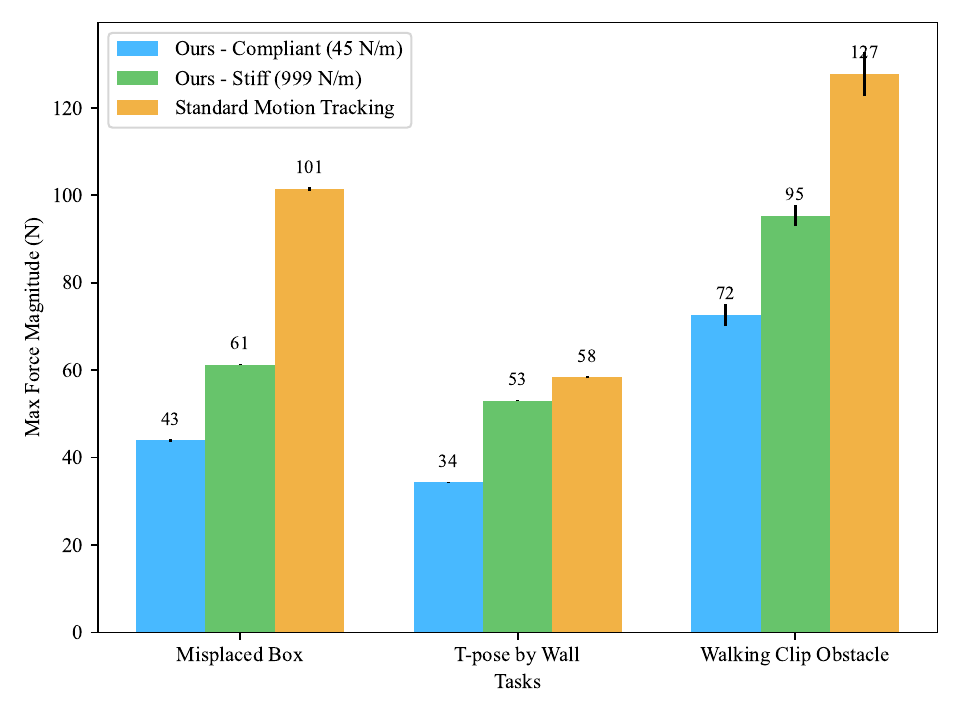}
    \includegraphics[clip,trim=5.5cm 3cm 5cm 13.23cm, width=0.9\columnwidth]{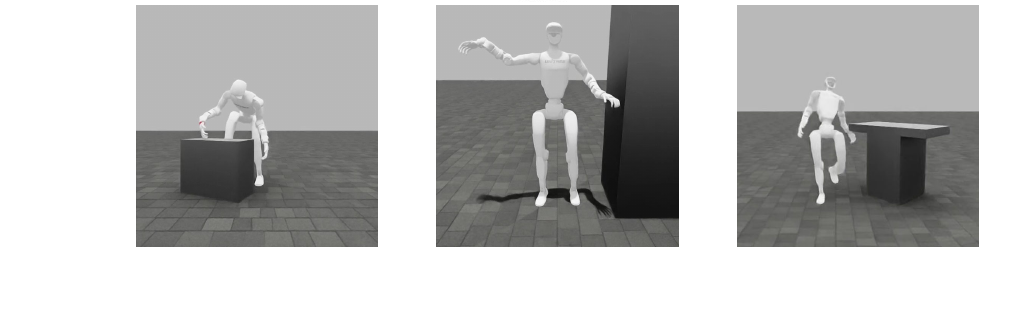}
    \caption{\textbf{SoftMimic reduces collision forces across various motions in unseen environments.} 
    The bar chart compares the maximum contact force generated by our policy (at low and high stiffness) and the stiff baseline across three challenging scenarios involving unexpected contact. In all cases, the compliant policy operating at a low stiffness significantly reduces interaction forces, enhancing safety. 
    }
    \label{fig:disturbance_rejection}
\end{figure}

\begin{figure*}[t]
    \centering
    \vspace{0.5em}
    \begin{tikzpicture}
    \node[inner sep=0, anchor=north] (base) {\includegraphics[width=0.9\linewidth]{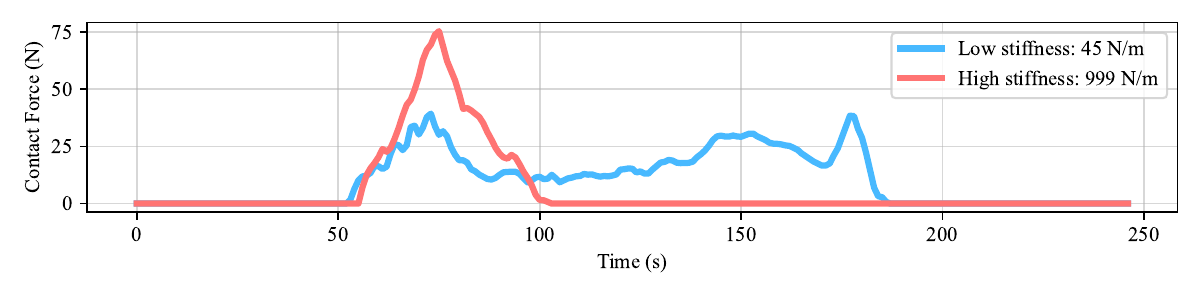}};
    \node[inner sep=0, anchor=north] at (base.north) {\includegraphics[clip,trim=0cm 0cm 0cm 4cm,width=0.9\linewidth]{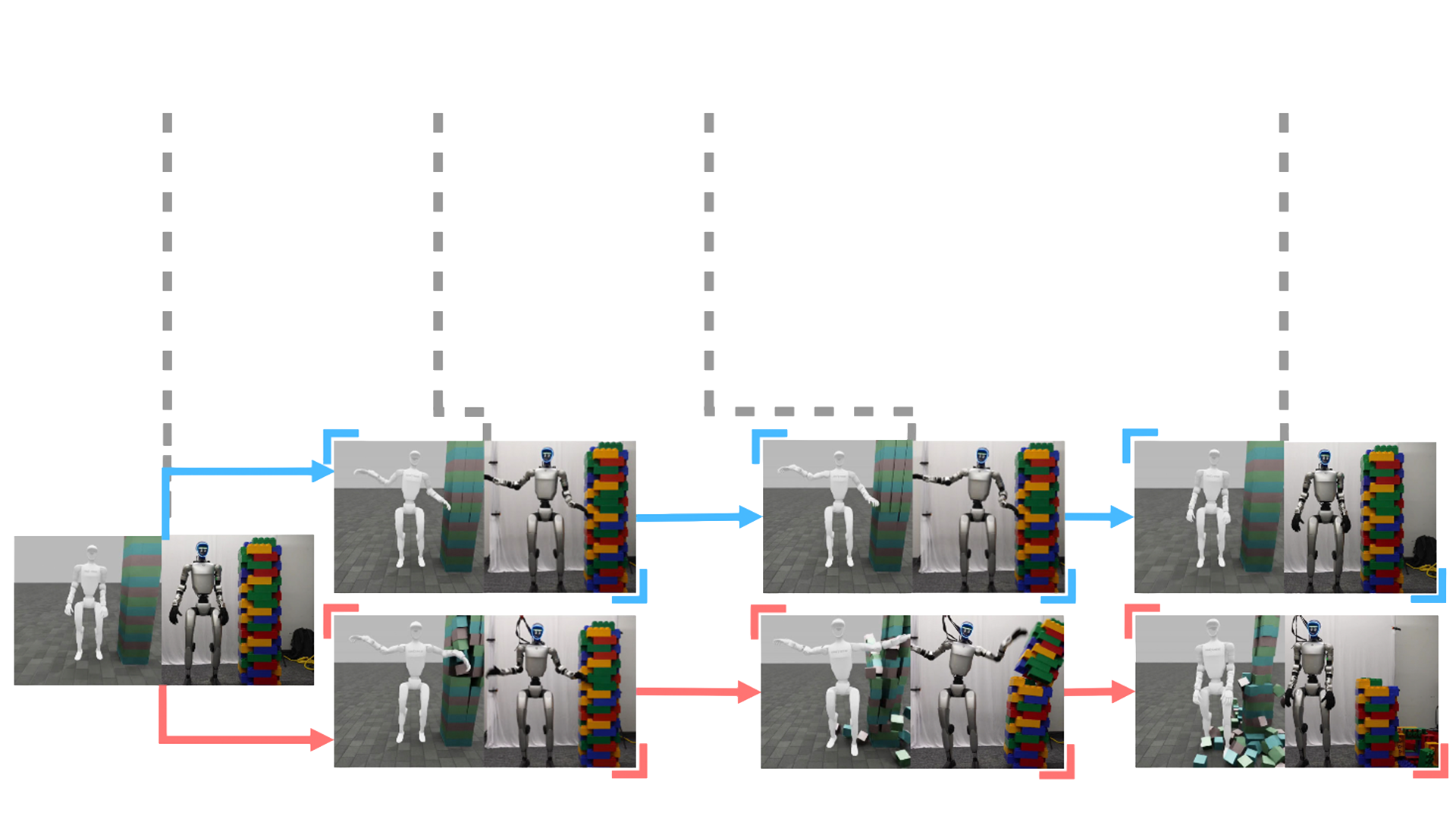}};
  \end{tikzpicture}
    \caption{\textbf{Stiffness modulation controls collision forces.} The plot shows the contact force over time as the robot's hand collides with a tower of blocks. By commanding different stiffness levels, our policy can produce low, controlled forces (blue) or high, potentially destructive forces (red), showcasing a direct trade-off between safety and posture tracking accuracy.}
    \label{fig:wall_collision}
\end{figure*}

\subsection{SoftMimic: Reinforcement Learning for CMT}
\label{sec:rl_compliance}

\textbf{Observation, Reward, Action Space.} We formulate compliant whole-body control as a reinforcement learning problem. The policy observes a state containing the robot's proprioceptive information $[q_t, \dot{q}_t]$, base state $[g^b_t, \omega^b_t]$, previous action $a_{t-1}$, and reference posture $q^{\text{ref}}_t$. The agent is rewarded with a sum of a DeepMimic-style reference motion tracking reward, $r_{\text{ref}} + r_{\text{smooth}}$, and a spring-like compliance reward, $r_{\text{spring}} = r_{\text{force}} + r_{\text{torque}} + r_{\text{pos}} + r_{\text{rot}}$, which depends on the current external wrench $W_i$. 
The policy outputs joint-space position targets for a PD controller with moderate gains, enabling torque control by modulating the position error.

\textbf{Observation Content.} The policy can implicitly learn admittance-style (estimate wrench, command pose), impedance-style (estimate pose, command wrench), or hybrid strategies depending on the desired stiffness and external force profile. Note that the policy directly observes neither the wrench nor displacement information, but can make inferences about them based on proprioceptive sensing. For an impedance strategy, the end-effector pose can be inferred from the joint positions $q_t$ and root orientation $g^b_t$ via forward kinematics. For an admittance strategy, the external wrench can be inferred from the robot's dynamics, using observations of previous joint position $q_{t-1}$, joint position target $a_{t-1}$, joint velocity $\dot{q}_{t-1}$, and joint accelerations (derived from $\dot{q}_t, \dot{q}_{t-1}, \omega^b_t, \omega^b_{t-1}$). To ensure this temporal information is available, the policy observes a history of the past 3 observation steps. As is standard in legged systems, the full root state and contact states are not directly observed; instead, the policy may partially infer them as needed, leveraging the associations between historical observations, commands, and simulation outcomes.

\begin{figure*}[t]
    \centering
    \vspace{0.5em}

    \begin{minipage}[b]{0.52\textwidth}
        \begin{subfigure}[b]{\linewidth}
            \includegraphics[width=0.9\linewidth]{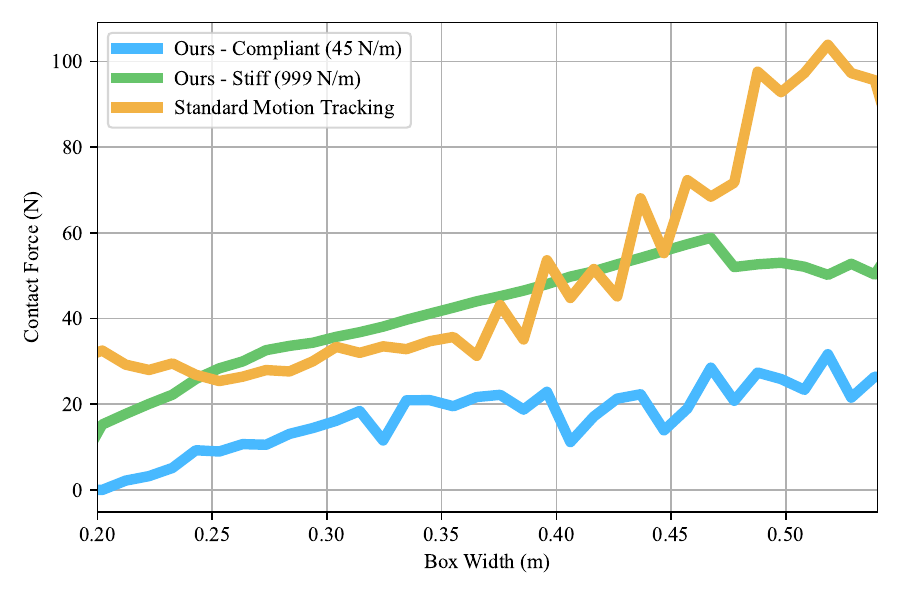}
            \includegraphics[clip,trim=5cm 3cm 5cm 12.6cm,width=\linewidth]{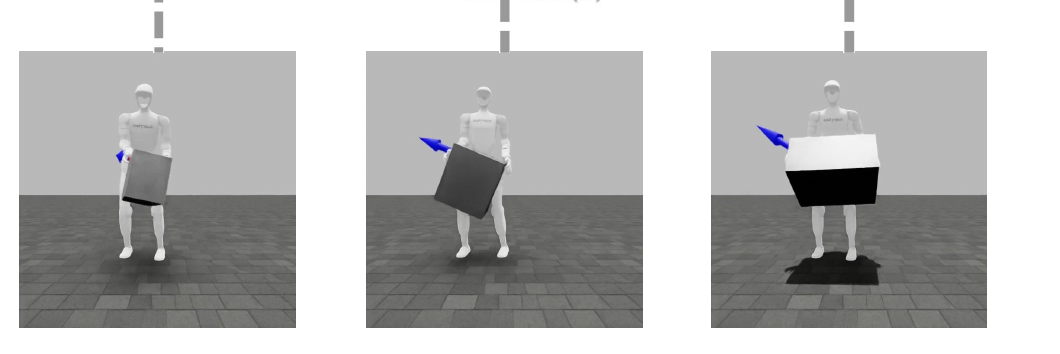}
            \caption{\textbf{Controllable interaction force:} Simulated normal contact force vs. box width using one object-agnostic reference motion. Softmimic force increases predictably with box width; a stiff tracker produces large spikes resulting in box damage or torque limit violations on real hardware.}
        \end{subfigure}
    \end{minipage}
    \hfill
    \begin{minipage}[b]{0.45\textwidth}
        \begin{subfigure}[b]{\linewidth}
            \includegraphics[width=\linewidth]{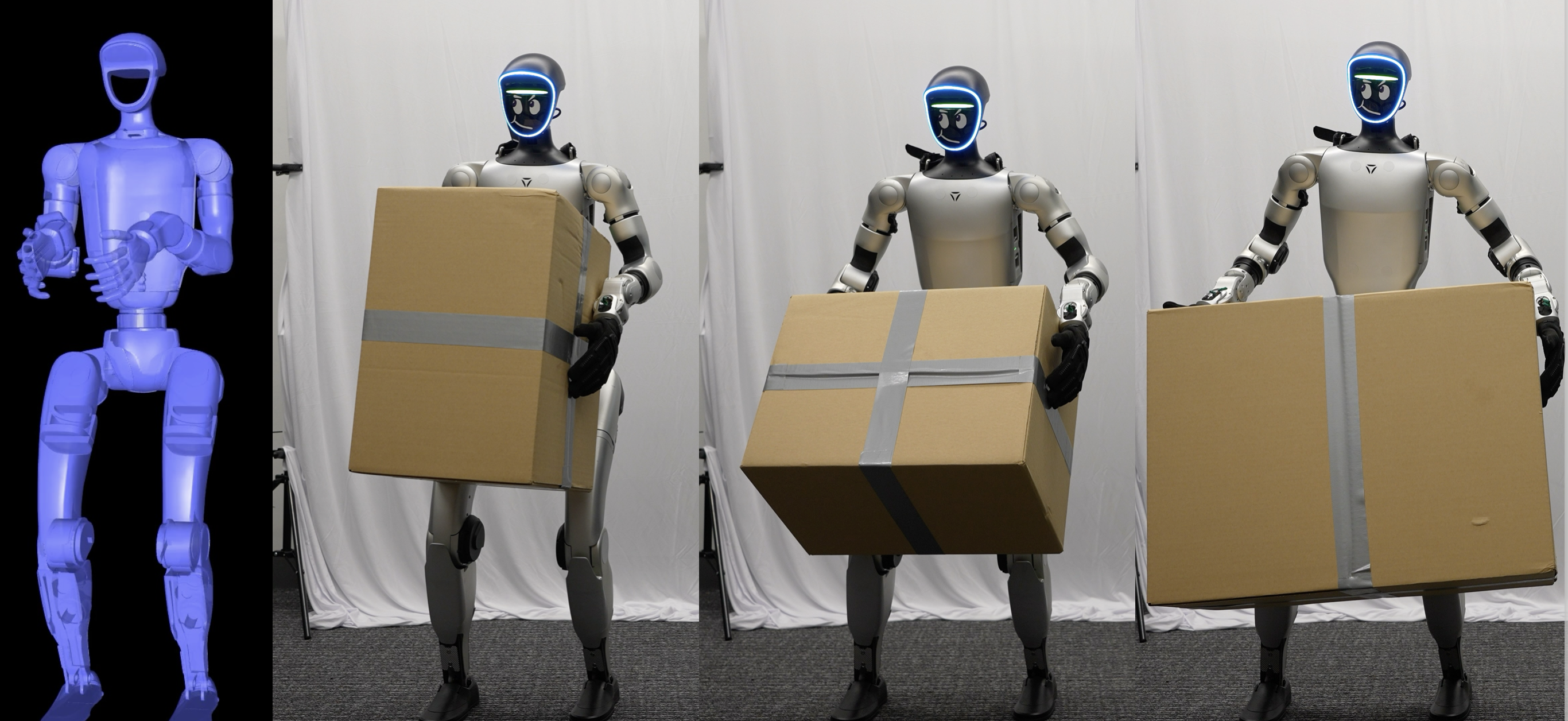}
            \caption{\textbf{Generalization of a single reference:} Real-world deployment with the same reference motion and stiffness: successful picks across multiple box widths with a gentle, consistent squeeze.}
        \end{subfigure}

        \vspace{0.8em} 

        \begin{subfigure}[b]{\linewidth}
            \includegraphics[width=\linewidth]{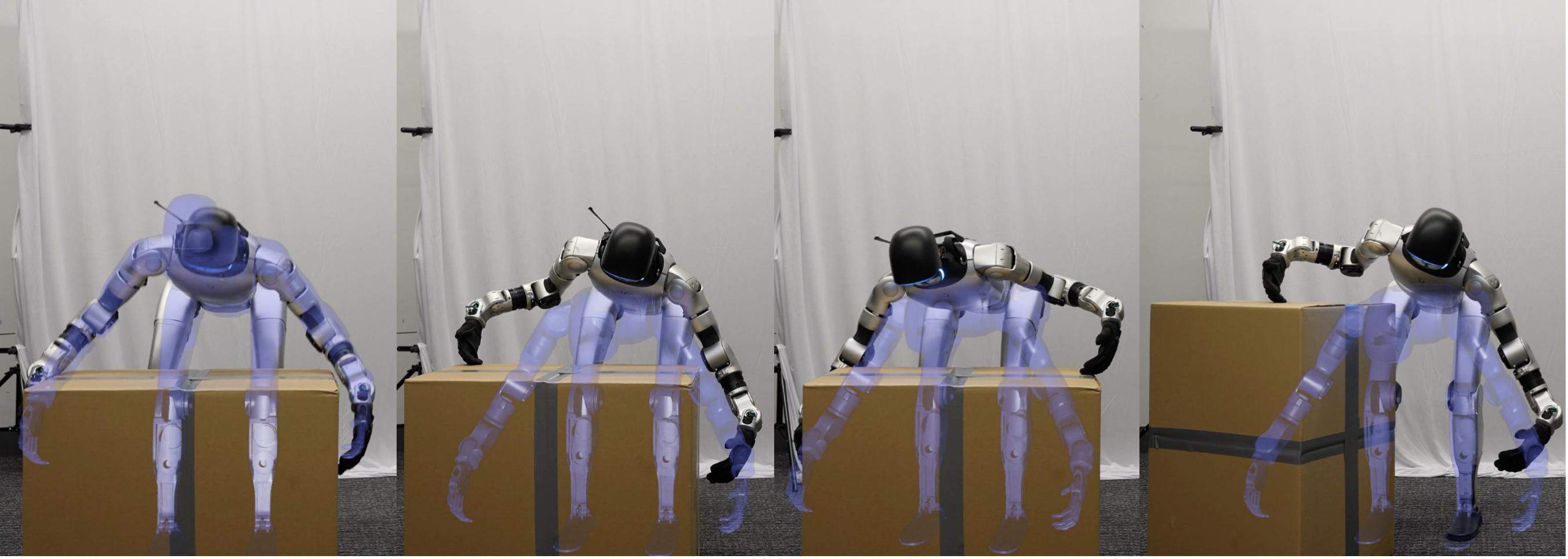}
            \caption{\textbf{Zero-shot robustness to misalignment:} picking a large box with nominal alignment (left) and under lateral/rotational misinitializations (middle–right). All behavior is achieved without simulating boxes or defining a prior over their location; robustness comes from SoftMimic training with generalized external forces.}
        \end{subfigure}
    \end{minipage}

    \caption{\textbf{SoftMimic enables generalization to unseen objects and disturbance scenarios.} Using a single motion reference designed for a 20cm box, our policy can successfully pick up boxes of increasing width. By commanding the same low stiffness, the robot not only successfully picks up different sized boxes with a consistent, gentle squeezing force, but it also is able to safely handle collisions with misaligned boxes. In contrast, the stiff baseline generates large and uncontrolled force spikes, risking damage to the object or robot.}
    \label{fig:task_generalization}
\end{figure*}

\textbf{Command Sampling and Force Field Dynamics.} Training episodes consist of sampling a motion clip, a desired robot stiffness, and an external force profile. The external force is implemented as a `force field'  \cite{portela2024learning} which pulls a selected link of the robot towards a moving setpoint with a distance-proportional force according to a randomized environment stiffness $K_{\text{env}}$ (Figure \ref{fig:system_diagram}). $K_{\text{env}} \to 0$ corresponds to a constant-force source (an admittance-like environment) and $K_{\text{env}} \to \infty$ corresponds to an immovable object (an impedance-like environment) \cite{ott2010unified}. Additional details of the force sampling parameters are provided in the appendix.

\begin{itemize}
    \item \textit{Stiffness Bounds:} When sensing and dynamics are noisy and the robot's state is only partially observable, inferences about pose and wrench also become noisy. This noise makes realizing highly sensitive responses, including extremely low or high stiffnesses, infeasible. To address this, we first train a state estimator to establish the approximate noise floor of the pose and wrench estimates. We then use these noise values in an idealized analysis to guide our stiffness sampling range. 
    We define requirements of $10\,\mathrm{N}$ force accuracy and $10\,\mathrm{cm}$ position accuracy, and empirically observe that the learned force estimator has average noise of $4\,\mathrm{N}$. 
    Thus, an admittance control strategy should be able to achieve the positioning target only for $K > \frac{4\,\mathrm{N}}{0.10\,\mathrm{m}} = 40\,\mathrm{N}/\mathrm{m}$. Likewise, with a position estimation noise of $1\,\mathrm{cm}$, an impedance controller can achieve the desired force accuracy only if $K < \frac{10\,\mathrm{N}}{0.01\,\mathrm{m}} = 1000\,\mathrm{N}/\mathrm{m}$. This analysis establishes approximate upper and lower feasible bounds for training. 
    \item \textit{Log-Uniform Sampling:} We aim to realize behaviors across a wide range of stiffness and compliance values. Since compliance is the inverse of stiffness, uniformly sampling stiffness would heavily bias the dataset towards high-stiffness, low-compliance behaviors, and vice versa. A change in stiffness from $1040$ to $1080\,\mathrm{N/m}$ is a minor tweak to a stiff behavior, while a change from $40$ to $80\,\mathrm{N/m}$ is a significant change for a compliant one. To ensure we explore these different regimes equally, we sample both the robot and environment stiffness from a log-uniform distribution. 
    \item \textit{Velocity-based Event Sampling:} Suppose that every point in space has the same probability of containing a stationary collision surface. Then if the robot is moving with no information about its surroundings, its probability of some point on the robot colliding is proportional to the point's velocity. Therefore, we sample force event onsets for each link with probability proportional to its velocity, with a small constant additional probability for colliding while stationary.
\end{itemize}

\textbf{Early Termination and Reference State Initialization.} It is common practice to exploit early termination and reference state initialization to accelerate and stabilize motion imitation \cite{peng2018deepmimic}. 
A key advantage of our data augmentation approach (Section \ref{sec:data_shaping}) is that we can appropriately terminate and initialize episodes while the robot is under load by using the augmented compliant posture $q_{\text{aug}}$ as the reference. Without this augmented data, there would be no way to initialize the robot consistently with active wrenches,

\subsection{Compliant Motion Augmentation (CMA)}
\label{sec:data_shaping}

A key challenge with the RL problem posed in Section \ref{sec:rl_compliance} is that the final compliant posture arises from a competition between the motion tracking and spring-behavior rewards. This complicates exploration and makes the resulting behavior difficult to tune and predict.

Our proposed solution is to pre-generate an augmented motion dataset, $D_{\text{aug}}$, that explicitly contains desired whole-body responses to force events. This offline process enables two key advantages: 1) we can reject infeasible commands before RL training using simple kinematic and dynamic checks, and 2) we can precisely specify the desired compliant behavior through a structured optimization. We generate $D_{\text{aug}}$ using a differential inverse kinematics (IK) solver.

\textit{Task Hierarchy:} The IK solver optimizes for the following objectives:
\begin{enumerate}
    \item \textbf{Compliant Interaction (high priority; $w=5.0$).}
    For the interacting link $i$, we define the desired pose via the commanded stiffness $(\Kcmdt,\Kcmdr)$ and the wrench $\wi=(\Fi,\taui)$:
    \[
      \pdes{i}=\pref{i}+\frac{1}{\Kcmdt}\,\Fi,\qquad
    \]
    \[
      \Rdes{i}=\Rref{i}\exp\!\big([\taui/\Kcmdr]_\times\big).
    \]
    penalizing $\|\mathbf{p}_i(q)-\pdes{i}\|^2$ and $\|\log(\Rdes{i}^\top R_i(q))^\vee\|^2$.
    We perturb a single link at a time (hands in this work).
    \item \textbf{Foot Placement (high priority; $w=2.5$).} High-weight link pose tasks ensure that stance feet remain consistent with the reference contact schedule.
    \item \textbf{CoM Stabilization (medium priority; $w=0.1$).} A Center of Pressure (CoP)-aware Center of Mass (CoM) task provides moment compensation while allowing necessary body shifts.
    \item \textbf{Keypoint Posture (low priority; $w=0.01$).} Moderate-weight pose tasks on key links (e.g., elbows, shoulders, torso) preserve the original motion's style.
    \item \textbf{Joint Posture (very low priority; $w=10^{-4}$).} A regularization task tethers all degrees of freedom towards the reference configuration $q_{\text{ref}}$ to resolve redundancy.
\end{enumerate}
This hierarchy yields a continuous and feasible adapted joint trajectory, $q_{\text{aug}}(t)$, that embodies the desired compliant response across various interaction scenarios. When the IK solver fails to find a solution for a given wrench, we rewind the motion clip and iteratively scale down the wrench, rejecting the event entirely if the wrench falls below the sensing noise floor.

\subsection{Motion Data, Training Details, Baselines}
\textbf{Motion Data.} We trained and deployed compliant whole-body control policies on a Unitree G1 humanoid---one policy for each motion clip: standing, T-pose-move, walk, box-pick, pour, and dance, using identical hyperparameters. The motion data comes from the AMASS~\cite{mahmood2019amass} and LAFAN1~\cite{harvey2020robust} datasets, retargeted using methods from prior work~\cite{he2024learning, xie2025kungfubot}.

For each motion clip, we generate augmented data by solving the aforementioned inverse kinematics problem 
using Mink~\cite{zakka10mink, carpentier2019pinocchio} and MuJoCo~\cite{todorov2012mujoco}. The offline process is highly efficient, allowing us to generate 40 minutes of augmented data for a one-minute clip in approximately one minute of wall-clock time when parallelized. This produces a dataset of tuples $\big(\qref,\ \wi,\ \Kcmd,\ \qaug\big)$ that defines all interaction events for training.

\textbf{Training Hyperparameters.} Linear stiffness commands ranged from \SI{40}{\newton\per\meter} to \SI{1000}{\newton\per\meter}; angular stiffness from \SI{0.1}{\newton\meter\per\radian} to \SI{10}{\newton\meter\per\radian}. We train using PPO with the default hyperparameters from the IsaacLab and \texttt{rsl\_rl} libraries~\cite{schwarke2025rsl}.

\textbf{Baselines.} To rigorously evaluate our method, we compare it against two carefully designed baselines that aim to isolate the different components of our framework.

\begin{enumerate}
    \item \textbf{Stiff Baseline:} To demonstrate the value of explicit compliance, we first compare against a high-performance baseline analogous to standard motion imitation methods~\cite{peng2018deepmimic}. This \textbf{Stiff Baseline} is trained with a reward function that only incentivizes rigid tracking of the original reference motion, $q_{\text{ref}}$. Crucially, to ensure a fair and direct comparison, this baseline is exposed to the exact same distribution of external force perturbations during training as our compliant policy. This setup tests the emergent behavior of a state-of-the-art tracking controller when faced with physical interactions it is not explicitly rewarded to handle.

    \item \textbf{\texttt{no-aug} Ablation:} To specifically isolate the contribution of our learning-from-example data generation strategy, we design an ablation called \textbf{\texttt{no-aug}}. This policy is trained with the same spring-like compliance reward, $r_{\text{spring}}$, as our full method, but it does not have access to the augmented dataset $D_{\text{aug}}$. Consequently, its motion tracking reward, reference state initializations, and termination conditions are all based on the original non-compliant reference, $q_{\text{ref}}$. This creates a significant learning challenge: successful compliance generates a large tracking error relative to $q_{\text{ref}}$, which would normally trigger an early termination and thus penalize the desired behavior. To create a meaningful and learnable task, we modify the termination condition for this ablation: an episode only terminates if the robot's state deviates significantly from $q_{\text{ref}}$ \textit{without} satisfying the compliant displacement objective. This necessary adjustment allows the policy to explore compliant behaviors without being immediately punished, enabling a fair evaluation of learning with reward shaping alone.
\end{enumerate}

\begin{figure}[t]
    \centering
    \vspace{0.8em}
    \begin{subfigure}[b]{0.3\columnwidth}
        \includegraphics[width=\linewidth]{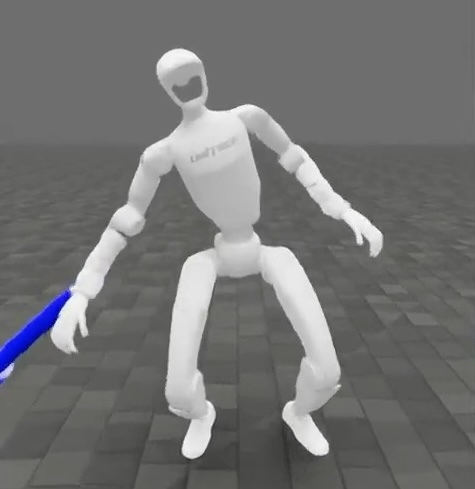}
        \includegraphics[width=\linewidth]{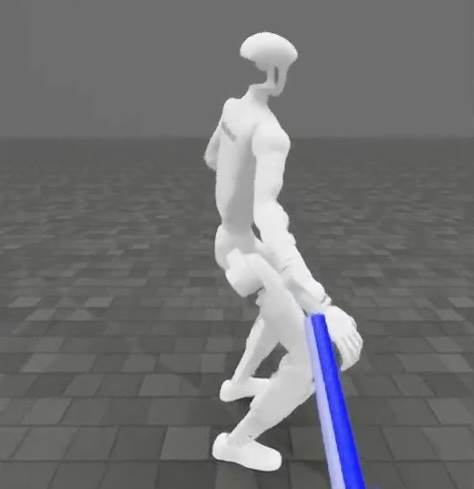}
        \caption{IK Style 1}
    \end{subfigure}
    \hfill
    \begin{subfigure}[b]{0.3\columnwidth}
        \includegraphics[width=\linewidth]{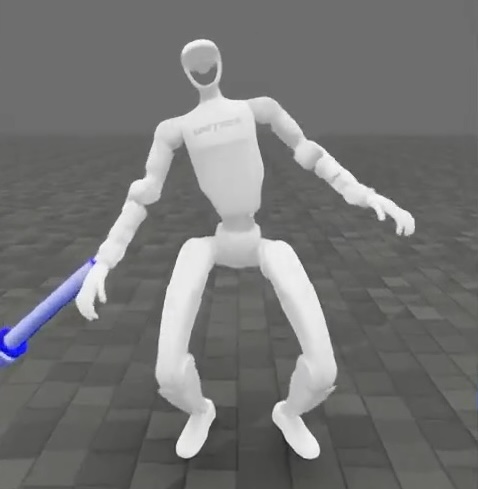}
        \includegraphics[width=\linewidth]{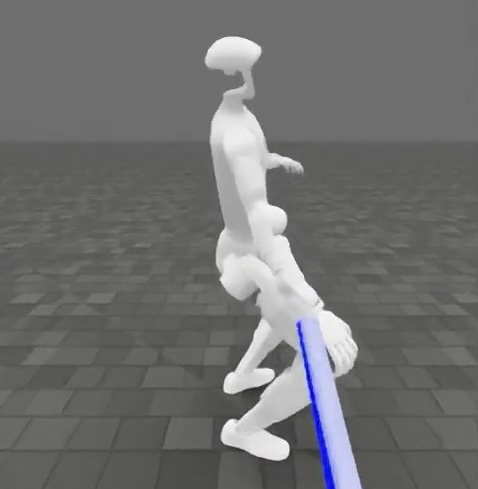}
        \caption{IK Style 2}
    \end{subfigure}
    \hfill
    \begin{subfigure}[b]{0.3\columnwidth}
        \includegraphics[width=\linewidth]{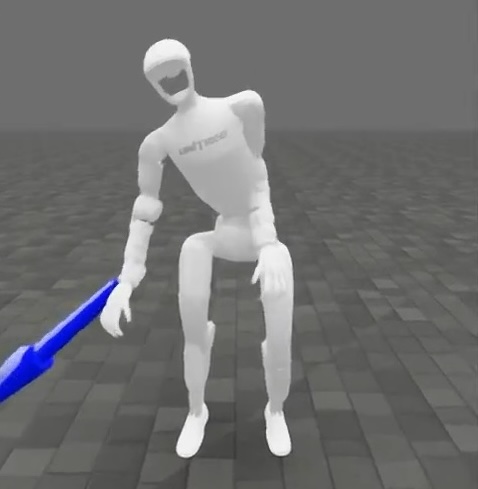}
        \includegraphics[width=\linewidth]{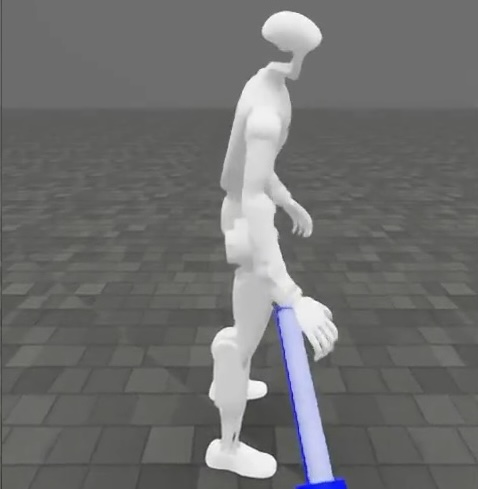}
        \caption{Ablation: \texttt{no-aug}}
    \end{subfigure}
    \caption{\textbf{Compliant Motion Augmentation provides fine-grained control over compliant style.} 
    Three different compliant policies are shown receiving the same external force in simulation with the same commanded stiffness. By adjusting cost terms in the offline IK solver—such as adding a pelvis orientation cost in (b) compared to (a)—we can specify distinct whole-body coordination strategies. The learned policies successfully reproduce the authored styles. In contrast, the policy trained without augmented data (\texttt{no-aug}, c) adopts an unpredictable emergent posture that also performs worse.
    }
    \label{fig:posture}
\end{figure}

\section{Results}
\label{sec:results}

\subsection{Motion Tracking Should Be Compliant}

\textbf{Compliance Improves Task Generalization.} Compliant imitation of a single motion can enable its generalization to different task variations. We demonstrate this through a box-picking motion. We apply SoftMimic with a single motion reference of a person picking a box of fixed size. During deployment, a natural approach compatible with non-compliant WBC would be to perceive the size and location of the box visually and map this to a reference motion -- either through an explicit perception module or via a learned high-level policy. We are interested in the scenario where the perception module is noisy and erroneously estimates the size or location of the box. In Figure \ref{fig:task_generalization}, we compare the force exerted in simulation on differently sized boxes by our compliant whole-body controller vs. the standard motion imitation approach; both are tracking a single original motion reference. Our method is able to maintain a lower squeezing force while successfully picking up differently sized boxes, while the standard method exhibits larger and unpredictable forces as it faces larger box sizes outside the scope of the original motion reference.

\textbf{Compliance Improves Disturbance Handling.} We evaluated the response of compliant policies to unseen environmental circumstances common during deployment of humanoid robots. 
\begin{itemize}
    \item \textit{T-Pose by Wall:} The robot attempts to raise its arm while standing next to a wall.
    \item \textit{Walking Clip Obstacle:} The robot walks past a table and its hand clips the corner.
    \item  \textit{Misplaced Box:} The robot attempts to execute a bending pick while the target box is not centered, and hits its hand on the top of the  box. 
\end{itemize}
 Figure \ref{fig:disturbance_rejection} reports the maximum force the robot exerts on the environment for each task, evaluated at two different stiffness levels of our method as well as with the standard motion tracking approach. The very compliant policy exerts significantly lower forces on the environment, showing that compliant whole body control can more safely handle forceful disturbances compared to existing baselines.

\textbf{Compliance Improves Safety.} Figure \ref{fig:wall_collision} shows how modulating the commanded stiffness results in drastically different environment interactions. A small stiffness results in the robot gently pushing on the tower and deviating significantly from the T-pose reference, while a large stiffness causes the robot to strongly resist deviations to the original reference motion, consequently exerting a large force on and toppling the blocks. 

\textbf{Sim-to-Real Validation.} Figure \ref{fig:setup} shows how the robot complies in the real world during various interactions, demonstrating generalization, disturbance handling, and safety. The useful behaviors associated with compliance all transfer to the real robot.

\begin{figure}[t]
    \centering
    \vspace{0.5em}
    \includegraphics[width=\columnwidth]{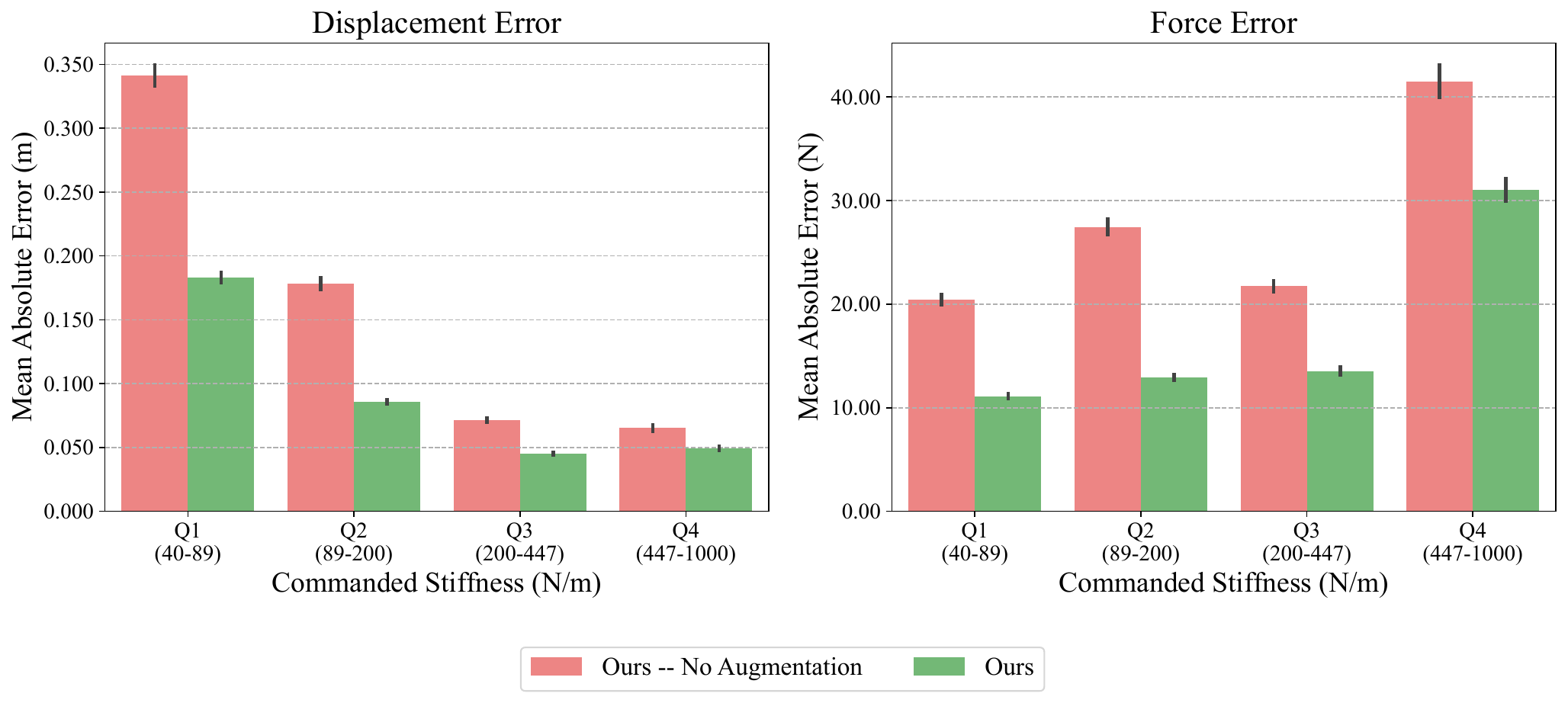}
    \caption{\textbf{Effect of Compliant Motion Augmentation on compliance accuracy.}
    Policies trained with augmented compliant trajectories (\textcolor{green}{green}) attain lower position and force error than the \texttt{no-aug} ablation (\textcolor{red}{red}), 
    with the largest gains at low stiffness where coordinated whole-body deviations are substantial.}
    \label{fig:aug_benefit}
\end{figure}

\subsection{Evaluating Stiffness Adherence}

We apply external forces to the standing robot in simulation and measure the resulting displacement across a range of stiffnesses. Figure \ref{fig:stiffness_adherence} shows the median effective stiffness (computed as the ratio between force and displacement) evaluated at various stiffness levels on a log-log plot. The standard motion tracking baseline, which is not conditioned on a stiffness command, yields an effective stiffness of about $500$.  As can be seen in the supplementary videos, the stiff policy preserves its posture when externally forced but tends to shuffle its feet, which registers as compliance in this evaluation conducted in the global reference frame. Our method displays a consistent sensitivity to the stiffness command across the entire range used in training. Figure \ref{fig:aug_benefit} shows the displacement is often regulated below \SI{10}{\centi\meter} and force error below \SI{15}{\newton} with exceptions at the lowest stiffnesses (elevated displacement error) and highest stiffnesses (elevated force error). Figure \ref{fig:stiffness_adherence} and \ref{fig:aug_benefit} also show that training with augmented references boosts performance compared to the ablation \texttt{no-aug}, particularly at low stiffnesses where it results in a 50\% reduction in displacement error.

\subsection{Data Shaping Controls Behavior}
\label{sec:ik_ablation}

A key benefit of our framework is the ability to resolve task specification ambiguities in the data augmentation stage. To illustrate this, we train compliant standing policies with two different IK-generated compliant datasets, one with a relatively higher pelvis orientation cost term that encourages the robot to squat and one with a relatively lower term that encourages the robot to bend. Figure \ref{fig:posture} shows how the resulting policies respond to perturbations in different styles depending on the behavior designed during the IK dataset generation. It also compares the behavior of the best \texttt{no-aug} policy, which displays an emergent postural response resulting from the balance of rewards, which cannot be predicted before performing the expensive RL training.

\subsection{Compliant Control Preserves Motion Quality}
\label{sec:long_clip}
Our proposed method achieves compliance when interacting with forces and preserves competitive motion tracking accuracy in the non-perturbed case, even for long and dynamic motion clips. Under no perturbations, we compare the joint position and keypoint tracking error of our method and standard motion tracking for skills used in our demonstrations, as well as a long, challenging dance clip (\texttt{dance1\_subject2} from LAFAN1 \cite{harvey2020robust}) which has recently been used to demonstrate the high performance of motion tracking systems. Table \ref{tab:tracking_comparison} shows both our compliant policy and the standard motion tracking baseline achieve small tracking errors. This minor increase in tracking error is an expected trade-off for learning a much richer and more versatile behavioral repertoire. Figure \ref{fig:all_motions_comparison} (Appendix) shows the training progression of total reward for SoftMimic vs. baseline and the convergence of compliance objectives. It takes a bit more time to train SoftMimic to convergence compared to stiff motion tracking. The policy must learn not only to track a single motion but also to embed a wide range of compliant responses.

\begin{table}[t]
    \centering
    \vspace{0.5em}
    \caption{\textbf{Motion tracking quality comparison.} Comparison of tracking error under no-perturbation conditions (free space) for our compliant policy and a stiff baseline on various skills. Errors are reported as mean joint position error (degrees) and keypoint Cartesian error (cm), with standard error of the mean over $36$ episodes. 
    }
    \label{tab:tracking_comparison}
    \setlength{\tabcolsep}{3.5pt} 
    \begin{tabular}{lcccc}
        \toprule
        \multicolumn{1}{c}{\bfseries Skill} & \multicolumn{2}{c}{\bfseries Ours (Compliant)} & \multicolumn{2}{c}{\bfseries Stiff Baseline} \\
        \cmidrule(lr){2-3} \cmidrule(lr){4-5}
         & Joint (°)& Keypoint (cm) & Joint (°) & Keypoint (cm) \\
        \midrule
        Box Pick  & $5.04 \pm 0.01$ & $2.65 \pm 0.01$ & $2.04 \pm 0.00$ & $1.36 \pm 0.00$ \\
        Walk      & $6.39 \pm 0.00$ & $3.44 \pm 0.00$ & $6.09 \pm 0.00$ & $3.50 \pm 0.00$ \\
        Dance     & $11.10 \pm 0.01$ & $6.05 \pm 0.01$ & $5.16 \pm 0.01$ & $3.01 \pm 0.00$ \\
        \bottomrule
    \end{tabular}
\end{table}

\section{Conclusion}
This work introduced a formulation for learning compliant whole-body motion tracking for humanoid robots. We demonstrate that our compliant policy outperforms the standard motion tracking baseline in generalization to unseen manipulation scenarios and in safety when handling disturbances. In qualitative experiments, we see that different user-commanded stiffness values can drastically change how the robot interacts with the environment—from gently pushing to toppling a tower of blocks—and with people during human-robot interaction. Quantitatively, we find that a compliant robot can generalize to unseen objects without applying excessive force, and when colliding with a disturbance, our policy applies nearly half the force of the standard baseline. Our policy also demonstrates good adherence to the commanded stiffness across a wide range of values and can comply across a broad workspace. Finally, under unperturbed conditions, our approach preserves tracking performance comparable to a state-of-the-art baseline across a variety of motions, including whole-body locomotion and manipulation.

Looking forward, a key area for future work is determining how to best select stiffness for a given task. While our experiments showcase the benefits of lower stiffness for safety and generalization--demonstrating that a single fixed value can be effective for diverse scenarios such as lifting a misplaced object--we anticipate that real-world deployments will require dynamically adjusting stiffness; for example, using higher stiffness to lift a heavy box versus lower stiffness to gently hand an object to a person. Further improvements could also come from training a foundational compliant whole-body controller capable of tracking large-scale motion datasets or live teleoperation. The success of such a model depends heavily on the quality of its training data. Although our kinematic data augmentation was sufficient to realize useful behaviors, its quality could be enhanced by incorporating dynamics to yield more physically plausible motions.  We could also explore and address gaps in workspace coverage caused by our rejection sampling approach. In particular, our constraint on the positions of the feet during compliant motion augmentation may be unnecessarily restrictive in scenarios that require foot contact switching to realize compliance. Alternatively, future research could explore data-driven distance metrics to guide behavior within the nullspace of the compliant interaction task.
Another extension would be to generate augmented data for simultaneous forces on multiple links, rather than a single link at a time; our approach was able to generalize to some multi-contact scenarios like box picking, but the behavior could be further assured and regularized with explicit training.
Finally, motivated by the many benefits of whole-body compliance, another promising direction is to define compliant behavior for wrenches on any link of the body, rather than only the wrists, in pursuit of fine-grained stiffness control across the robot's entire surface.








\bibliographystyle{IEEEtran}
\bibliography{IEEEabrv,IEEEfull}{}

\clearpage
\newpage

\appendix

\renewcommand{\thesubsection}{\Alph{subsection}}

\subsection{RL Hyperparameters}

We train our policies using Proximal Policy Optimization (PPO) \cite{schulman2017proximal} implemented in the \texttt{rsl\_rl} library \cite{schwarke2025rsl}. Hyperparameters for the PPO algorithm are detailed in Table \ref{tbl:ppo_hparams}. The policy network is an MLP with hidden layers [512, 512, 256, 128] and an ELU activation function. The critic network is an MLP with hidden layers [512, 512, 512, 512].

\subsection{Training Environment Details}

\textbf{Observation Space:} The policy observation space, summarized in Table \ref{tab:observation_space}, provides proprioceptive feedback, information about the original (non-compliant) reference motion, and the commanded stiffness.

\textbf{Domain Randomization:} To promote robust sim-to-real transfer, we employ standard domain randomization during training, covering both robot dynamics and observation noise. The randomization ranges are specified in Table \ref{tab:domain_rand}.

\textbf{Reward Function:} The total reward is a weighted sum of terms designed to encourage motion tracking, compliant interaction, and physically stable behavior. The complete reward function is detailed in Table \ref{tab:reward_table}.

\subsection{Compliant Motion Augmentation Details}

Our offline data generation process uses a combination of procedural event sampling and inverse kinematics to create a rich dataset of feasible compliant behaviors. The goal is to produce stylistically consistent whole-body motions that correctly respond to generalized external forces at various stiffnesses. The process can be broken down into three main stages: Event Generation, IK Solving, and Feasibility Validation.

\subsubsection{Event Generation}
We generate two distinct types of interaction events to ensure the policy learns a versatile set of responses. Each event is defined by a target link, a time profile, and interaction parameters. During RL training, both event types (ramp and collision) are simulated by the same virtual forcefield equation \cite{portela2024learning}. The only difference between modes is the final motion path of the forcefield's origin relative to the reference motion. 

\paragraph{Ramped Wrench Events} This event type is designed to simulate controlled interactions. The sampling process proceeds in the following order:
\begin{enumerate}
    \item \textbf{Timing and Link:} An event start time is chosen after a randomized rest period (see Table \ref{tab:combined_data_aug_ik_params}). A target link (left or right hand) is selected uniformly.
    \item \textbf{Stiffness Sampling:} The desired robot stiffness ($K_{\text{robot}}, k_r$) is sampled from a \textit{log-uniform} distribution. This ensures balanced coverage of both very compliant and very stiff behaviors.
    \item \textbf{Constrained Displacement Sampling:} Crucially, we do not sample force directly. Instead, we compute a valid range of Cartesian displacements for the target link. This range is constrained by the maximum allowed force and displacement limits, given the stiffness sampled in the previous step. A target displacement is then sampled uniformly from this valid range.
    \item \textbf{Peak Wrench Calculation:} The peak force $\mathbf{F}_{\text{ext}}$ is calculated as the product of the sampled stiffness and displacement. Its direction is sampled uniformly on a unit sphere. The same logic applies to the peak torque $\boldsymbol{\tau}_{\text{ext}}$.
    \item \textbf{Profile Timing:} A target interaction speed is sampled. This speed is used to calculate the event's ramp-up duration ($\|\text{displacement}\| / \|\text{speed}\|$), ensuring physically plausible motion. A hold duration is then sampled, defining the complete ramp-hold-ramp profile of the event.
\end{enumerate}

\paragraph{Simulated Collision Events} To better emulate unexpected contact, this mode spawns a virtual collision point in the path of the reference motion. The interaction force is generated organically from the penetration depth of the reference hand relative to this point, governed by the sampled robot and environment stiffness values. The probability of a collision point spawning at any timestep is set proportional to the link's velocity in the reference motion.

\subsubsection{IK Solving}
For each timestep of a generated event, we use a differential IK solver (Mink~\cite{zakka10mink} with DAQP) to find a full-body configuration $\mathbf{q}_{\text{aug}}$ that satisfies the compliant objective while maintaining balance and motion style. The solver minimizes a weighted sum of cost terms, where each term corresponds to a task objective as detailed in Table \ref{tab:combined_data_aug_ik_params}.

\subsubsection{Feasibility Validation and Rejection Sampling}
At every timestep during an event, the resulting IK solution is checked against a set of hard feasibility constraints (see Table \ref{tab:combined_data_aug_ik_params}). If any criterion is violated:
\begin{enumerate}
    \item The event's magnitude is scaled down by a factor (we use $0.8$). For ramped events, this means reducing the peak force; for collision events, this means shortening the event duration.
    \item The entire event is re-simulated from its start time with the reduced magnitude.
    \item This process repeats until the event is fully feasible or its magnitude falls below a minimum threshold (e.g., \SI{1}{\newton}), at which point the event is rejected and discarded from the dataset.
\end{enumerate}
This iterative rejection sampling is critical for ensuring that the final augmented dataset $D_{\text{aug}}$ contains only kinematically achievable and well-behaved compliant motions, simplifying the subsequent RL training problem.

\begin{figure*}[t!] 
    \centering
    \includegraphics[width=0.3\textwidth]{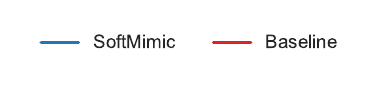}
    \vspace{2mm}
    
    \begin{subfigure}[b]{0.32\textwidth}
        \includegraphics[width=\linewidth]{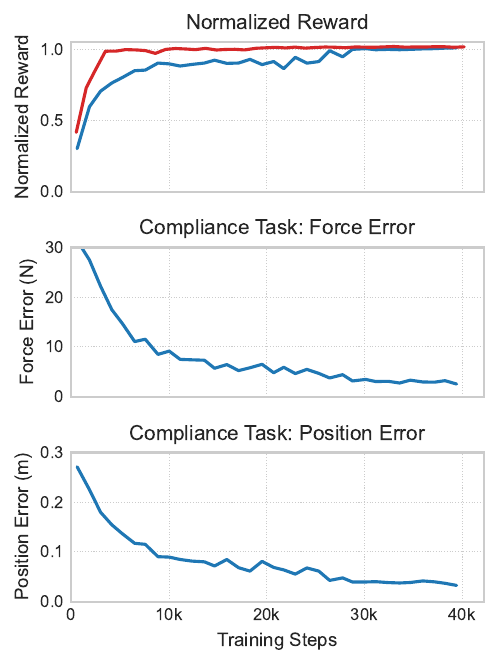}
        \caption{Stand}
        \label{subfig:stand}
    \end{subfigure}
    \hfill 
    \begin{subfigure}[b]{0.32\textwidth}
        \includegraphics[width=\linewidth]{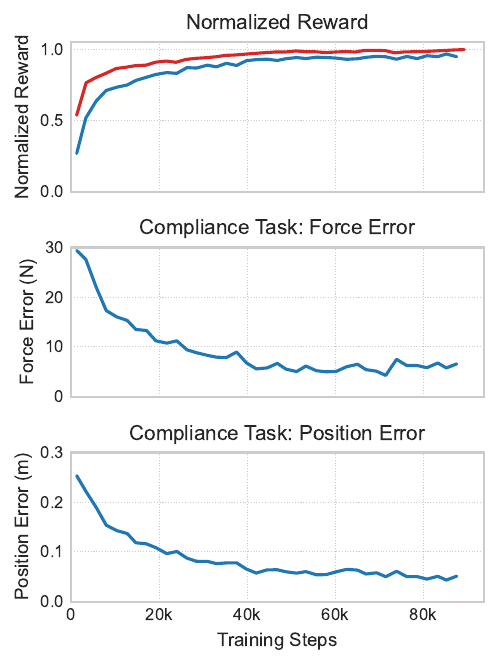}
        \caption{Box Pick}
        \label{subfig:boxpick}
    \end{subfigure}
    \hfill
    \begin{subfigure}[b]{0.32\textwidth}
        \includegraphics[width=\linewidth]{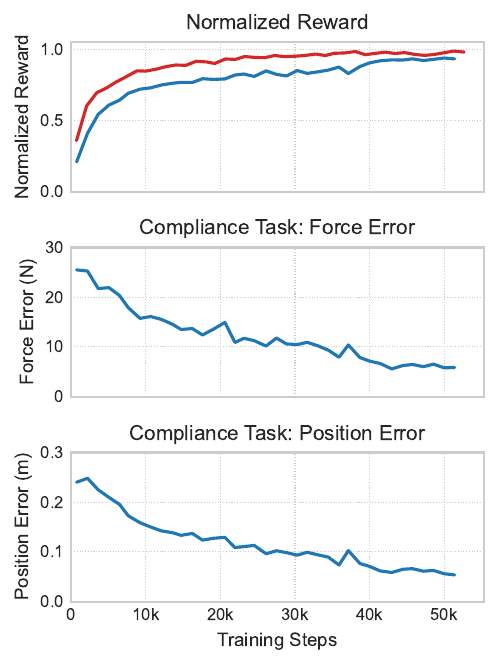}
        \caption{T-Pose}
        \label{subfig:tpose}
    \end{subfigure}
    
    \vspace{4mm} 
    
    \begin{subfigure}[b]{0.32\textwidth}
        \includegraphics[width=\linewidth]{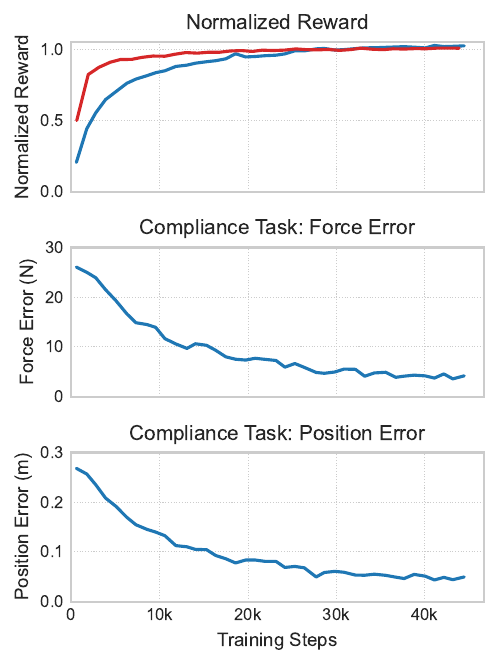}
        \caption{Pour}
        \label{subfig:pour}
    \end{subfigure}
    \hspace{1cm}
    \begin{subfigure}[b]{0.32\textwidth}
        \includegraphics[width=\linewidth]{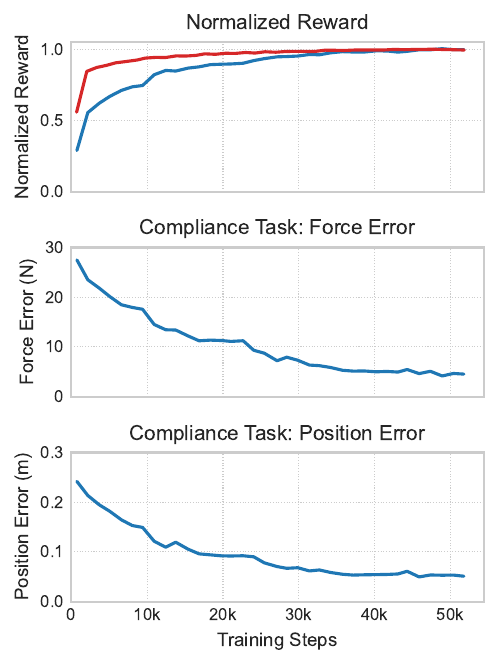}
        \caption{Walk}
        \label{subfig:walk}
    \end{subfigure}
    
    \caption{
        Training performance comparison between SoftMimic and a baseline across five distinct motions.
        The top plot for each motion shows the normalized reward, where both policies are compared.
        The middle and bottom plots show force and position imitation error, respectively, for only the SoftMimic policy to highlight the training progression of its compliant behavior. Training with SoftMimic incurs modestly slower convergence while the policy learns a rich set of responses for different stiffnesses.
    }
    \label{fig:all_motions_comparison}
\end{figure*}

\begin{table}
    \centering
    \bgroup
    \def\arraystretch{1.3}
    \caption{PPO hyperparameters.}
    \label{tbl:ppo_hparams}
    \scriptsize
    \begin{tabular}{lr}
        \toprule
        Hyperparameter & Value \\
        \midrule
        \# Environments & 4096 \\
        Timesteps per Rollout & 24 \\
        Discount Factor ($\gamma$) & 0.99 \\
        GAE Parameter ($\lambda$) & 0.95 \\
        Learning Rate & $1 \times 10^{-3}$ \\
        Schedule & Adaptive (KL target: 0.01) \\
        Epochs per Rollout & 5 \\
        Minibatches per Epoch & 4 \\
        Value Loss Coefficient & 1.0 \\
        Entropy Bonus & 0.002 \\
        Clip Range & 0.2 \\
        Max Gradient Norm & 1.0 \\
        Optimizer & Adam \\
        \bottomrule
    \end{tabular}
    \egroup
\end{table}

\begin{table}
    \centering
    \bgroup
    \def\arraystretch{1.3}
    \caption{Policy observation space.}
    \label{tab:observation_space}
    \scriptsize
    \begin{tabular}{p{0.3\linewidth}p{0.6\linewidth}}
        \toprule
        Component Group & Description \\
        \midrule
        \textbf{Proprioception} & Joint positions (relative to default), joint velocities, base angular velocity, and projected gravity vector. A history of the last 3 timesteps is included. \\
        \addlinespace
        \textbf{Reference Motion} & Reference joint positions, root height, gravity vector, base linear and angular velocity, and foot contact schedule. Includes current state, a history of 3 timesteps, and a future horizon of 20 points sampled up to 1 second ahead. \\
        \addlinespace
        \textbf{Task Commands} & Logarithm of the desired translational and rotational stiffness. A history of the last 3 timesteps is included. \\
        \addlinespace
        \textbf{Action History} & The previous action taken by the policy over the last 3 timesteps. \\
        \bottomrule
    \end{tabular}
    \egroup
\end{table}

\begin{table*}[t!]
    \centering
    \bgroup
    \def\arraystretch{1.5}
    \caption{\textbf{Comprehensive Data Augmentation and IK Solver Parameters.} This table details the three key components of our offline data generation pipeline. The IK solver minimizes a weighted sum of squared error norms, $\sum_i w_i \| \mathbf{e}_i \|^2$, where the error terms $\mathbf{e}_i$ are defined below. This optimization is subject to the data generation and feasibility parameters that govern the sampling of interaction events and the rejection of kinematically infeasible outcomes.}
    \label{tab:combined_data_aug_ik_params}
    \scriptsize
    \begin{tabular}{p{3.4cm} p{6.2cm} p{3.5cm} c}
        \toprule
        \bfseries Component & \bfseries Mathematical Formulation / Description & \bfseries Value / Range & \bfseries Units \\
        \midrule
        \multicolumn{4}{l}{\bfseries IK Solver Objective Function Terms ($w_i \| \mathbf{e}_i \|^2$)} \\
        \cmidrule(lr){1-4}
        Compliant Interaction &
        $\mathbf{e}_{\text{pos}} = \pdes{i} - \mathbf{p}_i(\mathbf{q})$ \newline
        $\mathbf{e}_{\text{rot}} = \log\!\left(\Rdes{i}^\top \mathbf{R}_i(\mathbf{q})\right)^\vee$
        & Weight ($w_i$): 5.0 & - \\
        \addlinespace
        Foot Placement &
        $\mathbf{e}_{\text{pos}} = \mathbf{p}_{\text{ref,foot}} - \mathbf{p}_{\text{foot}}(\mathbf{q})$ \newline
        $\mathbf{e}_{\text{rot}} = \text{log}(\mathbf{R}_{\text{ref,foot}}^T \mathbf{R}_{\text{foot}}(\mathbf{q}))^\vee$
        & Weight ($w_i$): 2.5 & - \\
        \addlinespace
        CoM Stabilization &
        $\mathbf{e}_{\text{CoM}} = \mathbf{c}_{\text{target}} - \mathbf{c}(\mathbf{q})$, where \newline 
        $\mathbf{c}_{\text{target},xy} = \mathbf{c}_{\text{ref},xy} + \frac{1}{Mg}[-\mathbf{m}_{\text{ext},y}, \mathbf{m}_{\text{ext},x}]$ \newline
        ($\mathbf{m}_{\text{ext}}$ is total moment about CoP)
        & Weight ($w_i$): 0.1 & - \\
        \addlinespace
        Keypoint Posture &
        Tracks Cartesian poses of key links (torso, elbows, knees) against the reference motion $\mathbf{q}_{\text{ref}}$.
        & Weight ($w_i$): 0.01 & - \\
        \addlinespace
        Joint Posture &
        $\mathbf{e}_{\text{joint}} = \mathbf{q}_{\text{ref}} - \mathbf{q}$
        & Weight ($w_i$): $10^{-4}$ & - \\
        \midrule
        \multicolumn{4}{l}{\bfseries Data Generation Hyperparameters} \\
        \cmidrule(lr){1-4}
        Robot Stiffness (Linear) & Log-uniform sampling of commanded robot stiffness $\Kcmdt$. & $[40, 1000]$ & \SI{}{\newton\per\meter} \\
        Robot Stiffness (Angular) & Log-uniform sampling of commanded robot stiffness $\Kcmdr$. & $[0.1, 10]$ & \SI{}{\newton\meter\per\radian} \\
        Environment Stiffness (Linear) & Stiffness of the virtual force field or collision plane $\Kenvt$. & $[10, 1000]$ & \SI{}{\newton\per\meter} \\
        Environment Stiffness (Angular) & Rotational stiffness of the virtual force field $\Kenvr$. & $[0.1, 10]$ & \SI{}{\newton\meter\per\radian} \\
        Peak Force Limit & Hard constraint on max peak force $\|\mathbf{F}_{\text{ext}}\|$. & $140$ & \SI{}{\newton} \\
        Peak Torque Limit & Hard constraint on max peak torque $\|\boldsymbol{\tau}_{\text{ext}}\|$. & $10$ & \SI{}{\newton\meter} \\
        Displacement Limit & Hard constraint on max resulting displacement. & $0.7$ & \SI{}{\meter} \\
        Ang. Displacement Limit & Hard constraint on max resulting angular displacement. & $2.0$ & \SI{}{\radian} \\
        Time Between Events & Randomized rest period between interaction events. & $[0.5, 1.5]$ & \SI{}{\second} \\
        Event Hold Duration & Duration of the peak force/torque application. & $[0.5, 1.0]$ & \SI{}{\second} \\
        Target Interaction Speed & Sampled velocity used to calculate force ramp duration. & $[0.1, 1.0]$ & \SI{}{\meter\per\second} \\
        \midrule
        \multicolumn{4}{l}{\bfseries Feasibility Rejection Criteria} \\
        \cmidrule(lr){1-4}
        Max Link Tracking Error & Maximum deviation of the solved hand pose from its compliant target pose $(\pdes{i},\Rdes{i})$. & Threshold: $0.05$ & \SI{}{\meter} \\
        Max Stance Foot Displacement & Maximum deviation of solved stance foot poses from their reference poses. & Threshold: $0.05$ & \SI{}{\meter} \\
        Max CoM Tracking Error & Maximum deviation of the solved CoM from its CoP-aware target in the XY-plane. & Threshold: $0.15$ & \SI{}{\meter} \\
        \bottomrule
    \end{tabular}
    \egroup
\end{table*}

\subsection{Training Convergence Details}
We render training curves for SoftMimic and the baseline in Figure \ref{fig:all_motions_comparison}. This illustrates the relative convergence speed of SoftMimic and the dynamics of learning force and displacement adherence over time. x-axis is number of policy update steps (each training step processing a 24-timestep rollout across 4096 parallel environments).

\begin{table}
    \centering
    \bgroup
    \def\arraystretch{1.3}
    \caption{Domain randomization ranges.}
    \label{tab:domain_rand}
    \scriptsize
    \begin{tabular}{lcr}
        \toprule
        Parameter & Range & Units \\
        \midrule
        \multicolumn{3}{l}{\textbf{Dynamics Randomization (per episode)}} \\
        \cmidrule(lr){1-3}
        Payload Mass (added to torso) & $[-2.0, 2.0]$ & \SI{}{\kilogram} \\
        Link Mass Scale & $[0.7, 1.3]$ & - \\
        Base CoM Displacement (XYZ) & $[-0.02, 0.02]$ & \SI{}{\meter} \\
        Joint Damping (added) & $[0, 2]$ & \SI{}{\newton\meter\second\per\radian} \\
        Joint Armature (added) & $[0.01, 0.1]$ & \SI{}{\kilogram\meter^2} \\
        Joint Friction (added) & $[0, 0.01]$ & \SI{}{\newton\meter} \\
        Ground Static/Dynamic Friction & $[0.5, 2.0]$ & - \\
        Ground Restitution & $[0.0, 0.5]$ & - \\
        \midrule
        \multicolumn{3}{l}{\textbf{Observation Noise (per step)}} \\
        \cmidrule(lr){1-3}
        Joint Position Noise & $[-0.01, 0.01]$ & \SI{}{\radian} \\
        Joint Velocity Noise & $[-1.5, 1.5]$ & \SI{}{\radian\per\second} \\
        Base Angular Velocity Noise & $[-0.2, 0.2]$ & \SI{}{\radian\per\second} \\
        Projected Gravity Noise & $[-0.01, 0.01]$ & - \\
        \bottomrule
    \end{tabular}
    \egroup
\end{table}

\begin{table}
    \centering
    \bgroup
    \def\arraystretch{1.0}
    \caption{Reward terms and weights.}
    \label{tab:reward_table}
    \scriptsize
    \begin{tabular}{p{0.8\linewidth}r}
        \toprule
        Term Description & Weight \\
        \midrule
        \multicolumn{2}{l}{\textbf{Compliance Rewards}} \\
        \cmidrule(lr){1-2}
        Force Link Position Tracking (vs. desired compliant $\pdes{i}$) & 3.0 \\
        Force Link Orientation Tracking (vs. desired compliant $\Rdes{i}$) & 3.0 \\
        Applied Force Tracking (vs. desired $\Fi$) & 2.0 \\
        Applied Torque Tracking (vs. desired $\taui$) & 2.0 \\
        \midrule
        \multicolumn{2}{l}{\textbf{Motion Tracking Rewards}} \\
        \cmidrule(lr){1-2}
        Keypoint Position Tracking (vs. augmented ref, from $\qaug$) & 2.0 \\
        Keypoint Orientation Tracking (vs. augmented ref, from $\qaug$) & 2.0 \\
        Base Orientation Tracking (vs. augmented ref, from $\qaug$) & 0.5 \\
        Base Linear Velocity Tracking (vs. augmented ref, from $\qaug$) & 0.5 \\
        Base Angular Velocity Tracking (vs. augmented ref, from $\qaug$) & 0.5 \\
        \midrule
        \multicolumn{2}{l}{\textbf{Stability and Regularization Rewards}} \\
        \cmidrule(lr){1-2}
        Alive & 1.5 \\
        Joint Position Limits Penalty & -10.0 \\
        Stance Foot Stability (sliding penalty) & -0.005 \\
        Joint Velocity L2 Penalty & -2.8e-4 \\
        Action Rate L2 Penalty & -0.01 \\
        Stance Foot Joint Motion Penalty & -0.4 \\
        \bottomrule
    \end{tabular}
    \egroup
\end{table}

\end{document}